\pdfoutput=1

\documentclass[11pt]{article}

\usepackage[final]{acl}

\usepackage{times}
\usepackage{latexsym}

\usepackage[T1]{fontenc}

\usepackage[utf8]{inputenc}

\usepackage{microtype}

\usepackage{inconsolata}

\usepackage{graphicx}
\usepackage{subcaption}

\usepackage{CJKutf8}

\usepackage{color}

\usepackage{multirow}

\title{Does Alignment Tuning Really Break LLMs' Internal Confidence?}
\author{
 \textbf{Hongseok Oh}
  \quad \textbf{Wonseok Hwang\thanks{Corresponding author}}
\\ University of Seoul
\\ \texttt{\{cxv0519, wonseok.hwang\}@uos.ac.kr}
}

\begin{document}
\begin{CJK}{UTF8}{mj}
\maketitle
\begin{abstract}

Large Language Models (LLMs) have shown remarkable progress, but their real-world application necessitates reliable calibration. This study conducts a comprehensive analysis of calibration degradation of LLMs across four dimensions: models, calibration metrics, tasks, and confidence extraction methods.
Initial analysis showed that the relationship between alignment and calibration is not always a trade-off, but under stricter analysis conditions, we found the alignment process consistently harms calibration.
This highlights the need for (1) a careful approach when measuring model confidences and calibration errors and (2) future research into algorithms that can help LLMs to achieve both instruction-following and calibration without sacrificing either.

\end{abstract}

\section{Introduction} \label{sec:intro}

The impressive performance of LLMs has opened up unprecedented possibilities for their application in real-world. Calibration, a measure of model reliability, is crucial in such applications \cite{ece, sce}.
Recently, there have even been cases of legal repercussions for using fabricated case law generated by large language models (LLMs) in legal documents \cite{lawyergpt}.
However, recent studies indicate that the alignment process, a necessary step to improve instruction-following capabilities of LLMs, harms their calibration \cite{gpt4,onthecalibration}.
Prior study \cite{justask} has explored the calibration of black-box models with respect to confidence extraction methods and calibration metrics, and they argued that the calibration of LLMs after alignment may not degrade the reliability depending on the confidence extraction method.
However, by focusing on the black-box API, they have not fully explored diverse logit-based confidence extraction methods.
We extend this work to open LLMs, by comprehensively analyzing the calibration degradation across models, metrics, tasks, and confidence extraction methods.

\begin{figure}[t]
\centering
\includegraphics[width=\columnwidth]{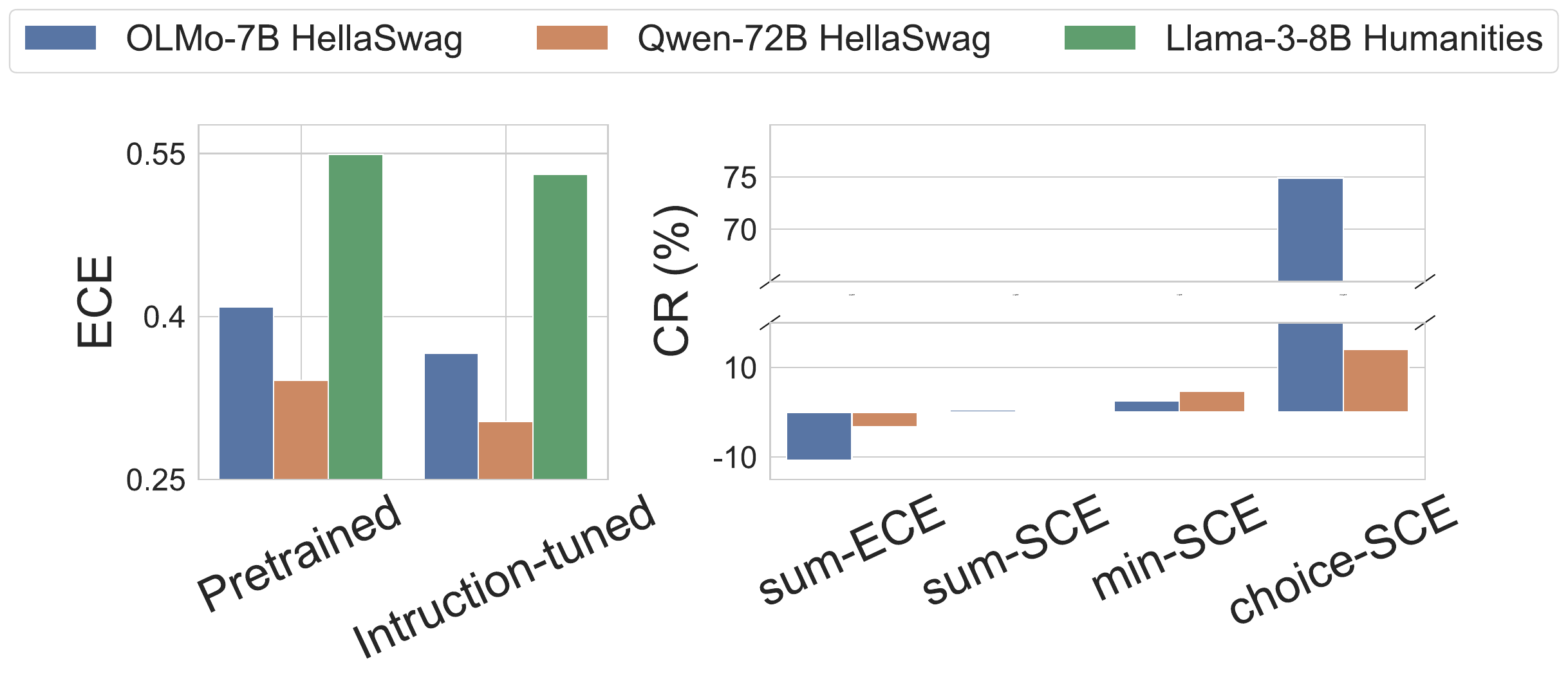}
\vspace{-2em}
\caption{Expected Calibration Error (ECE) scores of pretrained or instruction-tuned LLMs (left). The ECE change rates (CR) vary significantly depending on the choice of metrics (right).}
\label{fig:ecesce}
\vspace{-1em}
\end{figure}

\section{Experiment} \label{sec:exper}

We investigate calibration changes across 1) various open LLMs; Llama2 \cite{llama2}, Llama3 \cite{llama3}, Mistral \cite{mistral}, Gemma \cite{gemma}, OLMo \cite{olmo}, Qwen \cite{qwen}; 2) different metrics: Expected Calibration Error (ECE) \cite{ece} and Static Calibration Error (SCE) \cite{sce}; 3) various tasks: ARC-easy \cite{arc}, HellaSwag \cite{hellaswag}, MedMCQA \cite{MedMCQA}, MMLU \cite{MMLU} and PIQA \cite{PIQA}; 4) different methods to extract model confidence: \texttt{continuation-sum}, \texttt{continuation-min}, and \texttt{choice}. \texttt{continuation-sum} uses sum of logits for each choice's continuation token sequence given a context to extract confidence of model. We noticed that choices often have similar content (e.g., ["2.2x1011 kg", "2.2x1014kg", "2.2x1020kg", and "2.2x1016 kg"] in MMLU), so we introduce \texttt{continuation-min}, which uses the minimum logit within a continuation sequence, to mitigate the influence of this similarity and emphasize differences. \texttt{choice} method uses the logit of the token corresponding to the capital letter of each choice in the prompt. This eliminates any overlap in the choices' content. We follow \citet{bayesian} to design prompt for the choice method when continuation is the default.
All tasks were performed in a zero-shot setting using lm-evaluation-harness \cite{eval-harness} with four A6000 GPUs.

\begin{figure}[t]
\centering
\includegraphics[width=\columnwidth]{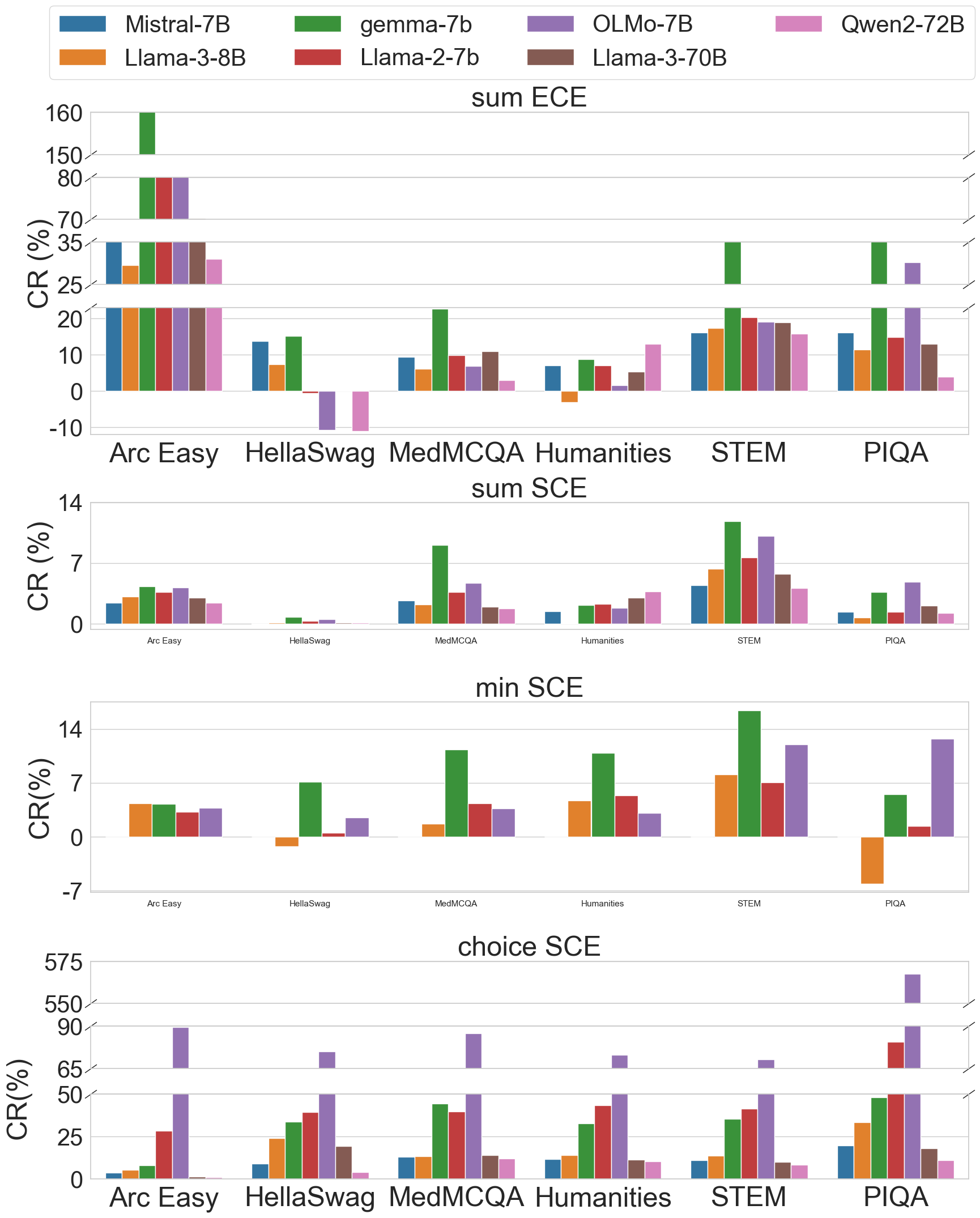}
\vspace{-2em}
\caption{The rates of change in calibration metrics between instruction-tuned and pre-trained models.
\texttt{sum}, \texttt{min}, and \texttt{choice} correspond to \texttt{continuation-sum}, \texttt{continuation-min}, and \texttt{choice} methods, respectively. Humanities and STEM are from MMLU}
\label{fig:cra}
\vspace{-1em}
\end{figure}

\section{Results and Analysis}

We have identified model-task combinations where the ECE decreased after the alignment process (Figure \ref{fig:ecesce}, left), which contradicts previous studies \cite{gpt4,onthecalibration}. Examples of these model-task combinations include OLMO-7B on HellaSwag (blue), Qwen-72B on HellaSwag (orange), and Llama-3-8B on MMLU Humanities (green).
However, on the same combinations, when examining the SCE, we observed consistent increases or negligible changes in calibration error. Furthermore, when we changed the method for extracting the model's confidence, the rate of change increased significantly, as depicted in the right side of Figure \ref{fig:ecesce}. Given the consistent increase of SCE with the alignment process in the \texttt{continuation-sum} method, we decided to focus on the SCE for the remainder of our experiments.

Next, we analyzed \texttt{continuation-min} method.
Unlike the consistent increase of SCE observed in continuation-sum method, as shown in the third graph of Figure \ref{fig:cra}, we observed SCE decreases for some combinations in this method.
Specifically, the Llama-3-8B model on HellaSwag and PIQA tasks, and the Mistral-7B model on HellaSwag and PIQA tasks, showed SCE changes of -1.25\%, -2.67\%, -6.08\%, and -6.28\%, respectively.

Finally, we examined the case of extracting model confidence using the \texttt{choice} method.
As shown in the fourth graph of Figure \ref{fig:cra}, the choice method consistently exhibited positive SCE changes across all tasks and model combinations.
This shows that if we combining strict calibration metric and confidence extraction method, calibration, as influenced by the alignment process, is consistently harmed regardless of the task or model.
Furthermore, as seen in Table \ref{table:rank}, the ranking of models' change rates remained remarkably robust across tasks.
Since the models we considered do not have significant structural differences, we believe that the consistent model ranking in Table \ref{table:rank} is primarily attributable to the training data and algorithms used in the model's alignment process.
Further experimental results are available in the following GitHub repository. \url{https://github.com/abzb1/alingment_calibration}

\begin{table}
\centering
\tiny
\centering
\caption{Rank of the change rate of the SCE for the task and models, when extracting confidence using the \texttt{choice}. Higher ranks assigned to larger changes.}
\vspace{-1em}
\begin{tabular}{c|c|c|c}
\hline
\multirow{2}{*}{Rank} & ARC\_Easy, MMLU\_STEM,  & HellaSwag & MedMCQA \\
 & MMLU\_Humanities, PIQA & & \\
\hline
1 & OLMo-7B & OLMo-7B & OLMo-7B \\
2 & Llama-2-7b & Llama-2-7b & gemma-7b \\
3 & gemma-7b & gemma-7b & Llama-2-7b \\
4 & Llama-3-8B & Llama-3-8B & Llama-3-70B \\
5 & Mistral-7B & Llama-3-70B & Llama-3-8B \\
6 & Llama-3-70B & Mistral-7B & Mistral-7B \\
7 & Qwen2-72B & Qwen2-72B & Qwen2-72B \\
\hline
\end{tabular}
\label{table:rank}
\vspace{-1.5em}
\end{table}

\section{Discussion and Conclusion}

Our analysis has yielded several insights.
Firstly, The calibration of LLMs can vary complexly and diversely depending on the combinations we've examined, so careful analysis is required.
Secondly, in the \texttt{choice}-SCE combination, where redundance of choices is minimized and confidence of all choices are considered, all models exhibited positive calibration error change rates across all tasks. This implies that under stricter analysis conditions, the alignment process can still be seen as detrimental to calibration.

\appendix

\bibliography{custom}

\end{CJK}
\end{document}